%% file: cvpr.tex
\documentclass[final]{cvpr}

\usepackage{times}
\usepackage{epsfig}
\usepackage{graphicx}
\usepackage{amsmath}
\usepackage{amssymb}
\usepackage{booktabs}
\usepackage{makecell}
\usepackage{multirow}
\usepackage{paralist} % compactenum
\usepackage{rotating}
\usepackage{colortbl}
\usepackage[dvipsnames]{xcolor}
\usepackage[ruled]{algorithm2e}

\usepackage{pifont}
\newcommand{\cmark}{\ding{51}}%
\newcommand{\xmark}{\ding{55}}%

\usepackage{appendix}

\newcommand{\sref}[1]{\S\ref{#1}}

\usepackage[pagebackref=true,breaklinks=true,colorlinks,bookmarks=false]{hyperref}

\def\cvprPaperID{1292} % *** Enter the CVPR Paper ID here
\def\confYear{CVPR 2021}
\makeatletter
\newcommand{\printfnsymbol}[1]{%
        \textsuperscript{\@fnsymbol{#1}}%
}
\makeatother

\begin{document}

%%%%%%%%% TITLE
\title{Learned Initializations for Optimizing Coordinate-Based Neural Representations}

% \author{
% Matthew Tancik\thanks{Authors contributed equally to this work.}\\
% \and
% Ben Mildenhall\printfnsymbol{1}\\
% \and
% Terrance Wang\\
% \and
% Divi Schmidt\\
% \and
% Pratul P. Srinivasan\\
% \and
% Jonathan T. Barron\\
% \and
% Ren Ng\\

% {$^1$UC Berkeley \quad $^2$Google Research}

% }

% \newcommand{\newand}{\quad\quad\,\,}
% \author{
% Matthew Tancik$^1$\thanks{Authors contributed equally to this work.}$^1$
% \newand
% Ben Mildenhall\printfnsymbol{1}$^1$
% \newand
% Terrance Wang$^1$
% \newand
% Divi Schmidt$^1$\\
% Pratul P. Srinivasan$^2$
% \newand
% Jonathan T. Barron$^2$
% \newand
% Ren Ng$^1$ \\
% $^1$UC Berkeley \newand $^2$Google Research
% }

\author{
Matthew Tancik$^{*1}$
\qquad
Ben Mildenhall$^{*1}$
\qquad
Terrance Wang$^1$
\qquad
Divi Schmidt$^1$\\
Pratul P. Srinivasan$^2$
\qquad
Jonathan T. Barron$^2$
\qquad
Ren Ng$^1$ \\
$^1$UC Berkeley \qquad $^2$Google Research
}

\maketitle

%%%%%%%%% ABSTRACT
\begin{abstract}
Coordinate-based neural representations have shown significant promise as an alternative to discrete, array-based representations for complex low dimensional signals. 
However, optimizing a coordinate-based network from randomly initialized weights for each new signal is inefficient.
We propose applying standard meta-learning algorithms to learn the initial weight parameters for these fully-connected networks based on the underlying class of signals being represented (e.g., images of faces or 3D models of chairs). 
Despite requiring only a minor change in implementation, using these learned initial weights enables faster convergence during optimization and can serve as a strong prior over the signal class being modeled, resulting in better generalization when only partial observations of a given signal are available.
We explore these benefits across a variety of tasks, including representing 2D images, reconstructing CT scans, and recovering 3D shapes and scenes from 2D image observations. 
\end{abstract}

%%%%%%%%% BODY TEXT
\section{Introduction}

\begin{figure}[t]
\begin{center}
   \includegraphics[width=\linewidth]{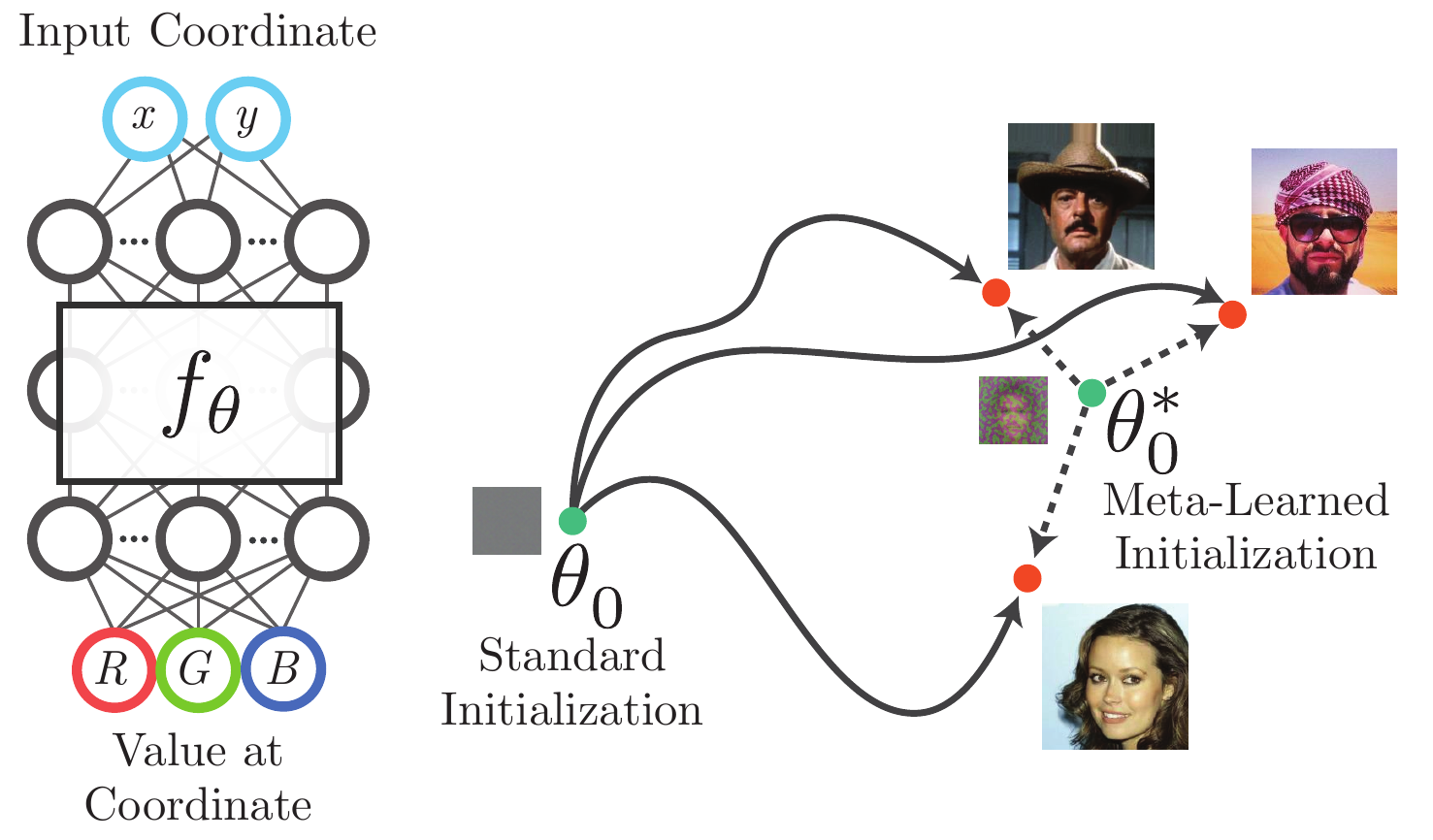}
\end{center}
   \caption{A coordinate-based MLP, illustrated on the left, takes a coordinate as input and outputs a value at that location. For example, the network could take in a pixel coordinate $(x,y)$ and emit the $(R, G, B)$ color at that pixel as output, thereby representing a 2D image. The network weights $\theta$ are typically optimized via gradient descent to produce the desired image, as depicted on the right. However, finding good parameters can be computationally expensive, and the full optimization process must be repeated for each new target. We propose using meta-learning to find initial network weights $\theta_0^*$ that allow for faster convergence and better generalization.}
\label{fig:teaser}
\end{figure}

\let\thefootnote\relax\footnote{* Authors contributed equally to this work.}

Recent work has demonstrated the potential of representing complex low-dimensional signals using deep fully-connected neural networks (typically referred to as multilayer perceptrons, or MLPs). A coordinate-based neural representation $f_\theta$ for a given signal is an MLP (with weights $\theta$) that is optimized to map from an input coordinate $\mathbf x$ to the signal's value at that coordinate. For example, $f_\theta$ could map from 2D pixel coordinates to RGB color values to encode an image. Unlike a signal stored as a discretely sampled array of values, a coordinate-based neural representation is \emph{continuous} and is not constrained to have a fixed spatial resolution. This fact has recently been exploited to design representations for 3D shapes (which typically occupy a small 2D subset of 3D space) that do not require cubic storage complexity, in contrast to 3D voxel grids~\cite{mescheder2019occ,nerf,park2019deepsdf,sitzmann2019srn}. 

However, one limitation of these neural representations is that computing network weights $\theta$ that reproduce a given signal typically requires solving an optimization problem by running many steps of gradient descent. This can take between seconds (when encoding a small image) and hours (when solving an inverse problem to recover a high resolution radiance field, as in NeRF~\cite{nerf}). 
Common approaches to address this issue include concatenating a latent vector to the input coordinate and supervising a single neural network to represent an entire class of signals~\cite{mescheder2019occ,park2019deepsdf}, or training a hypernetwork to map from signal observations (or a latent code) to MLP weights~\cite{sitzmann2020siren,sitzmann2019srn}. However, each of these strategies is restricted to representing only signals within its learned latent space, potentially limiting its ability to express previously unseen target signals.

Recent work~\cite{sitzmann2020metasdf} has shown that optimization-based meta-learning can dramatically reduce the number of gradient descent steps required to optimize a neural representation to encode a new signal in the case of signed distance fields of 2D and 3D shapes.
In this work, we propose learning the weight initialization for neural representations across a wide variety of underlying signal types, such as images, volumetric data, and 3D scenes.
We show that compared to a standard random initialization, using fixed, learned values for the initial network weights acts as a strong prior that enables both faster convergence during optimization and better generalization when only partial observations of the target signal are available. In the context of using neural representations for 3D reconstruction from images, a learned initialization specialized to a particular ShapeNet~\cite{shapenet2015} class allows the network to recover 3D shape from a single image over the course of optimization, whereas a standard randomly initialized network fails unless provided with multiple input views. Given a meta-training set consisting of observations of different signals sampled from a fixed underlying class, our setup applies an optimization-based meta-learning algorithm (MAML~\cite{maml} or Reptile~\cite{reptile}) in order to produce initial weights better suited for representing that specific signal class (e.g., face images from CelebA~\cite{liu2015celeba} or 3D chairs from ShapeNet~\cite{shapenet2015}).

The biggest advantage of our approach is its simplicity. Given an existing framework for test-time optimization of a neural representation, implementing an outer loop with MAML or Reptile update steps only requires a few extra lines of code and a dataset of training examples. Once the meta-learning phase is complete, the learned initial weights can be stored and later reloaded in place of a standard network initialization whenever a new signal needs to be encoded. This minor implementation change can significantly alter the behavior of the network during optimization. 
% For example, learning an initialization enabling a network optimized on only a single image to recover 3D structure.

%-------------------------------------------------------------------------
\section{Related Work}

\paragraph{Neural Representations}

Neural representations have recently risen to prominence as compact representations for 3D shapes. 
These methods represent shapes as implicit surfaces defined as a level set of an MLP network and enable full object reconstruction from incomplete 3D point cloud data or depth scans~\cite{implicitfields,neuralarticulated,learningshape,localdeep,localimplicit,mescheder2019occ,implicitlayers,park2019deepsdf}.
Later work combined this idea with various formulations of differentiable rendering to recover neural representations of 3D shape using only 2D image observations~\cite{implicitwithout3d,dist,nerf,niemeyer2020dvr,sitzmann2019srn,yariv2020idr}.

Coordinate-based neural networks have also been used to represent other low-dimensional signals, such as 2D images, where such networks (when trained via genetic algorithms) have been referred to as compositional pattern–producing networks~\cite{stanley2007cppn}. Recent works have shown that standard ReLU MLPs fail to adequately represent fine details in these complex low-dimensional signals due to a spectral bias~\cite{rahaman2019spectral} and address this issue by either replacing the ReLU activations with sine functions~\cite{sitzmann2020siren} or by lifting the input coordinates into a Fourier feature space~\cite{tancik2020fourfeat}. Our work makes use of these observations and presents a technique that enables a coordinate-based MLP to learn from the process of fitting many signals within a category so that it can quickly optimize to fit any new signal using fewer steps and fewer observations.

\paragraph{Meta-learning} 

Meta-learning typically addresses the problem of few-shot learning, where some examples of a given task (including training and test data) are used to learn an algorithm that achieves better performance on new, previously unseen instances of the same task. A prototypical example from computer vision is few-shot image classification, where a network must learn to differentiate between new classes at test time based on only a small number of labeled instances of each class.

Most relevant to this work are optimization-based meta-learning algorithms such as Model-Agnostic Meta Learning (MAML)~\cite{maml} and Reptile~\cite{reptile}, as well as various extensions~\cite{antoniou2018train,fallah2020convergence,flennerhag2019meta,li2017metasgd,imaml}. Given a network architecture for performing a task, these methods use an outer loop of gradient-based learning to find a weight initialization that allows the network to more efficiently optimize for new instances of the underlying task at test time. 
These methods assume the use of a standard gradient-based optimization method such as stochastic gradient descent or Adam~\cite{adam} at test time, making them easy to layer on top of existing implementations, as opposed to more complex methods such as Ravi \etal~\cite{ravi2017optfewshot}, which trains a ``meta-learner'' LSTM network to perform gradient updates for the underlying task. An exhaustive review of meta-learning algorithms is provided in the survey paper by Hospedales \etal~\cite{hospedales2020meta}. 

MetaSDF~\cite{sitzmann2020metasdf} specifically applies this idea of learning a weight initialization to the task of fitting neural representations to represent signed distance fields, and shows that this strategy achieves much more rapid convergence than standard approaches such as DeepSDF~\cite{park2019deepsdf}. 
Our work applies meta-learning to neural representations for a wider variety of underlying signal types and further explores the power of using initial weight settings as a prior.

%-------------------------------------------------------------------------
\section{Overview}

We define a finite signal $T$ as a function mapping from a bounded set $C\in \mathbb R^d$ to $\mathbb R^n$, where we refer to elements $\mathbf x \in C$ as $d$-dimensional coordinates. Examples include images (mapping from 2D pixel coordinates to 3D color values) or volumetric representations for 3D shapes (mapping from 3D locations to 4D tuples of color and density). A coordinate-based neural representation $f_\theta$ for $T$ is a fully connected neural network with $d$ input and $n$ output channels whose weights $\theta$ are optimized such that $f_\theta$ matches $T$ as closely as possible for all coordinates in $\mathbf x \in C$.

If direct pointwise observations $\{(\mathbf x_i, T(\mathbf x_i)\}_i$ of the signal $T$ are available, $f_\theta$ can be supervised by gradient descent using a simple L2 loss: 
\begin{equation}
    L(\theta) = \sum_i \| f_\theta(\mathbf x_i) - T(\mathbf x_i)\|_2^2 \, .
\end{equation}
Let $\theta_0$ denote the initial network weights before any gradient steps are taken, and let $\theta_i$ denote the weights after $i$ steps of optimization. Basic gradient descent applies the rule:
\begin{equation}
    \theta_{i+1} = \theta_i - \alpha \nabla_\theta L(\theta) |_{\theta=\theta_i} \, ,
\end{equation}
with a learning rate parameter $\alpha$, whereas more sophisticated optimizers such as Adam~\cite{adam} keep track of gradient moments over time to redirect the optimization trajectory. Given a fixed budget of $m$ optimization steps, different initial weight values $\theta_0$ will result in different final weights $\theta_m$ and signal approximation error $L(\theta_m)$. When emphasizing the functional dependence of $\theta_m$ on the initial weights and a particular signal, we will write $\theta_m(\theta_0, T)$.

It is often the case that only indirect observations of $T$ are available, taken through some forward measurement model $M(T, \mathbf p)$. For example, if $T$ is a 3D object, $M(T, \mathbf p)$ could be a 2D image captured of the object from camera pose $\mathbf p$. In this case, recovering a neural representation for $T$ from observations $\{\mathbf p_i, M(T, \mathbf p_i)\}_i $ requires solving an inverse problem by taking gradient steps on a loss that incorporates the forward model $M$:
\begin{equation}
    L_M(\theta) = \sum_i \|M(f_\theta, \mathbf p_i) - M(T, \mathbf p_i) \|_2^2 \, .
\end{equation}
If $M$ discards too much information about $T$ or the set of provided observations is too small, the resulting network $f_\theta$ may not match $T$ closely. For example, accurately recovering a 3D object from a single 2D view may not be possible without strong a priori knowledge of the object's shape.

\subsection{Optimizing initial weights}

We assume that we are given a dataset of observations of signals $T$ from a particular distribution $\mathcal T$ (e.g., 2D face images or 3D chairs) and our goal is to find initial weights $\theta_0^*$ that will result in the lowest possible final loss $L(\theta_m)$ when optimizing a network $f_\theta$ to represent a new, previously unseen signal from the same distribution:
\begin{equation}
    \theta_0^* = \mathrm{arg\, min}_{\theta_0} E_{T\sim \mathcal T}[L(\theta_m(\theta_0, T))]
\end{equation}
This problem of trying to learn the initial weights of a network to serve as a good starting point for gradient descent across a distribution of tasks is addressed by a variety of optimization-based meta-learning algorithms, such as MAML~\cite{maml} and Reptile~\cite{reptile}.

\paragraph{MAML~\cite{maml}} Given a task $T$, calculating the weight values $\theta_m(\theta_0, T)$ requires taking $m$ optimization steps, which are collectively referred to as the \emph{inner loop}. MAML wraps an \emph{outer loop} of meta-learning around this inner loop in order to learn the initial weights $\theta_0$. Each outer loop samples a signal $T_j$ from $\mathcal T$ and applies the update rule:
\begin{equation}
    \theta_0^{j+1} = \theta_0^j - \beta \nabla_{\theta} L(\theta_m(\theta, T_j))|_{\theta=\theta_0^j} 
\end{equation}
with meta-learning step size $\beta$. This update rule applies gradient descent to the loss on the weights $\theta_m(\theta_0^j, T_j)$ resulting from the inner loop optimization.

\paragraph{Reptile~\cite{reptile}} Reptile uses the same meta-learning setup as MAML but applies a simpler update rule that does not require calculating second-order gradients:
\begin{equation}
    \theta_0^{j+1} = \theta_0^j - \beta (\theta_m(\theta_0^j, T_j) - \theta_0^j) \, .
\end{equation}
This rule moves the previous weight initialization $\theta_0^j$ in the direction of the task-optimized weights $\theta_m(\theta_0^j, T_j)$.

\subsection{Experimental setup}

The meta-learning algorithms described previously are conceptually simple, requiring no changes to the architecture or optimization procedure of a coordinate-based neural representation when given a new signal to encode at ``test time'' (after meta-learning is complete). These algorithms produce only a set of initial network weights $\theta_0^*$ that are then used as a starting point for gradient descent. Test-time optimization on new signals is not limited to the same number of steps $m$ as were used in the inner loop during meta-learning; indeed, at test time we often observe benefits from optimizing for significantly more iterations than were used during the inner loop of the meta-learning algorithm.

MAML is typically able to produce a better initialization than Reptile given a fixed number of inner loop steps $m$, but Reptile can be unrolled for more inner loop steps because it is less memory-intensive than MAML. For some tasks, MAML's limited number of inner loop steps means that it can only observe a small percentage of the observations of a target signal. In these cases, we use Reptile to maximize the number of different observations seen over the course of the inner loop. Experimentally we find it beneficial to unroll more steps for more complex tasks.

Each of our experiments involves two phases:
\begin{compactenum}
    \item \emph{Meta-learning}, where we use MAML or Reptile in combination with a training dataset of example tasks (observations of different signal instances) to optimize initial network weights for that class of signals, and 
    \item \emph{Test-time optimization}, where we use standard gradient-based optimization to fit the weights of a network to observations of a previously unseen signal from the same class.
\end{compactenum}
We aim to answer the following question: how do different initial network weight settings influence the ability of a neural representation to fit to a new signal during test-time optimization?

%-------------------------------------------------------------------------
\section{Results}

We present results on 2D image regression, 2D computed tomography (CT) reconstruction, 3D object reconstruction, and 3D scene reconstruction. For each task, we demonstrate the benefits of using meta-learned initial weights optimized to reconstruct a specific class of signals. 
% We demonstrate the strength of this prior in various ways, comparing convergence speed, accuracy of the reconstruction, and ability to generate missing parts of the underlying signal given only partial observations.

For 2D image regression, a meta-learned weight initialization leads to faster convergence and better performance during test-time optimization.
For CT reconstruction, it allows for better reconstruction quality from fewer supervision views during test-time optimization.
For 3D shape reconstruction from images, it allows for faster convergence at test time and makes single view reconstruction possible.
For Phototourism landmark reconstruction, it can be optimized at test time to transfer the appearance of a single input image onto the whole landmark, which can then be rendered from novel camera views.

\subsection{Tasks}
\label{sec:tasks}
Here we provide the basic setup for each task. Please see the supplement for full implementation details.

\paragraph{Image regression} 
A prototypical example of a coordinate-based neural representation is an MLP optimized to represent a 2D image \cite{sitzmann2020siren,tancik2020fourfeat} by taking in 2D pixel coordinates and outputting RGB color values. We consider four different distributions $\mathcal T$: images of faces (\textit{CelebA}~\cite{liu2015celeba}), natural images (\textit{Imagenette}~\cite{imagenette}), images of text (\textit{Text}), and 2D signed distance fields of simple curves (\textit{SDF}). Each category contains around ten thousand examples. Given a sampled image $T\sim \mathcal T$, we provide all $178 \times 178$ pixels as observations for optimizing the network weights $\theta$ in the inner loop. 
Since this task is not memory constrained, we use MAML to meta-learn the weights over 2 unrolled gradient steps (separately for each category $\mathcal T$). In each of these inner loop steps, the entire image is reconstructed and used to calculate the loss.
% We use MAML to meta-learn the weights over 2 unrolled gradient steps (separately for each category $\mathcal T$) since this task is not memory constrained and can use full-batch gradient descent. 
For the MLP $f_\theta$, we use 5 layers with 256 channels each and sine function nonlinearities, as in SIREN~\cite{sitzmann2019srn}.

\paragraph{CT reconstruction} 
Computed tomography (CT) is a widely used medical imaging technique that captures projective measurements of the volumetric density of a target object. Tancik~\etal~\cite{tancik2020fourfeat} use a coordinate-based neural representation to reconstruct a 2D signal from 1D integral projections; the underlying MLP $f_\theta$ takes in a 2D coordinate and outputs a scalar volume density at that location. Here $\mathcal T$ is a dataset of 2048 randomly generated $256\times256$ pixel Shepp-Logan phantoms~\cite{shepp}, where we provide 2D integral projections of a bundle of 256 parallel rays from a random angle as the measurement for each sampled signal $T$ during meta-learning. We use Reptile to meta-learn the initial weights over 12 unrolled gradient steps. We found this to outperform MAML, which was limited to 3 unrolled steps due to memory constraints.
For the MLP $f_\theta$, we use 5 layers with 256 channels each and ReLU nonlinearities, and we apply random Fourier features to the input coordinates~\cite{tancik2020fourfeat}.

\begin{figure}[t]
\begin{center}
   \includegraphics[width=\linewidth]{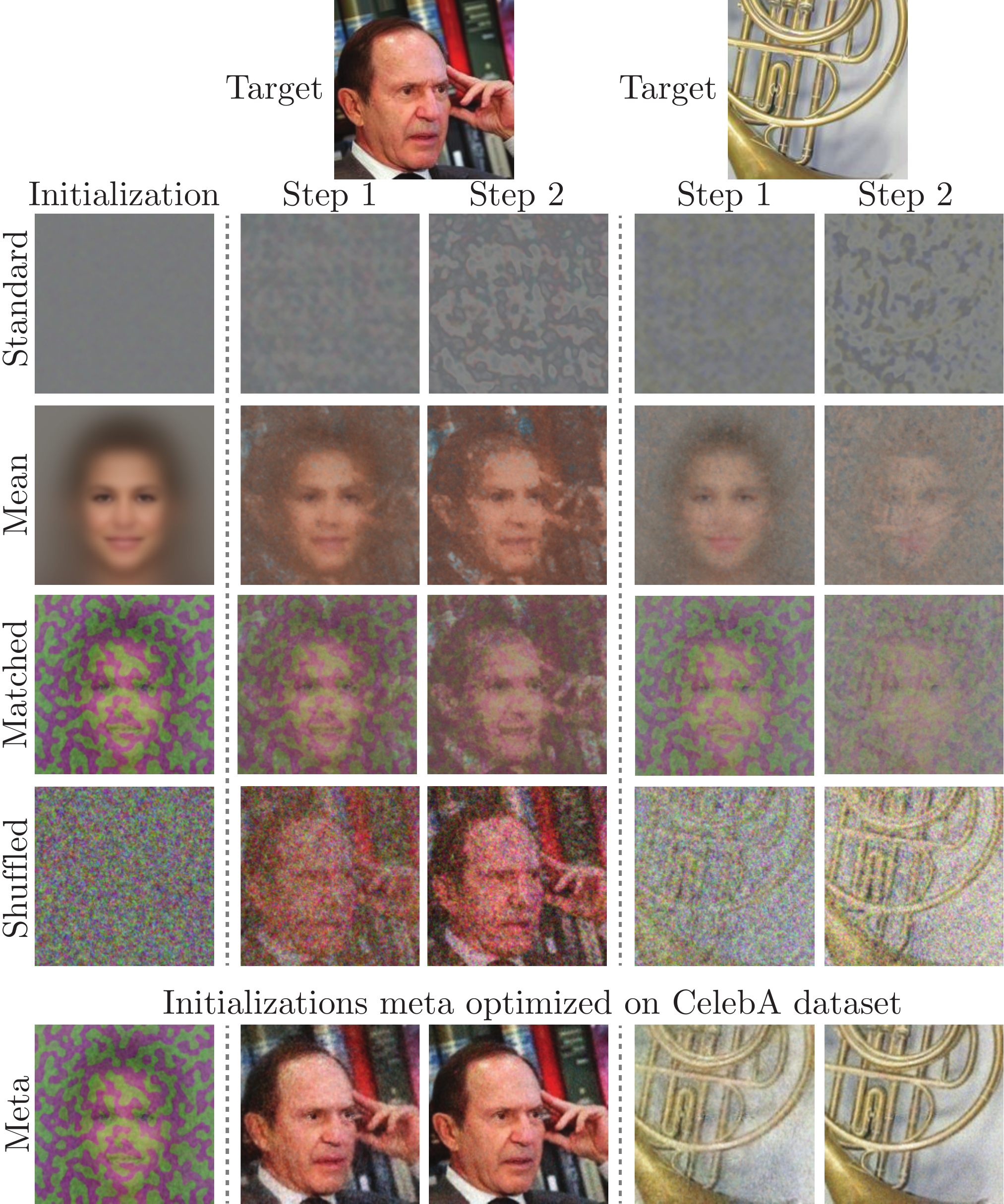}
\end{center}
   \caption{
   \textbf{Faster convergence:} Examples of optimizing a network to represent a 2D image from different initial weight settings. The meta-learned initialization (\textit{Meta}) is specialized for the class of human face images but still helps speed up convergence on other natural images (right). Non-meta-initialized networks take $10$ to $20$ times as many iterations to reach the same quality as the meta-initialized network does after only 2 gradient steps (see Table~\ref{tab:2d_image_mem}).}
\label{fig:2d_image_mem_all}
\end{figure}

\paragraph{View synthesis for ShapeNet~\cite{shapenet2015} objects} 
The goal of view synthesis is to generate a novel view of a scene from a set of reference images. Recently, neural radiance fields (NeRF)~\cite{nerf} proposed a method to accomplish this task by using a neural representation that predicts a color and density for any input 3D location and 2D viewing direction within the scene, along with a differentiable volumetric rendering model to generate new views from that representation. 
This network is optimized to minimize the residual of re-rendering each of the input reference images from their respective camera poses.
In our view synthesis experiments, we use a simplified NeRF model (simple-NeRF) that maintains the same image supervision and volume rendering context. Unlike the original NeRF model, we do not feed in the viewing direction and we use a single model instead of the two ``coarse'' and ``fine'' models used by NeRF.

For view synthesis on objects from the ShapeNet~\cite{shapenet2015} dataset, we consider three categories $\mathcal T$: \textit{Chairs}, \textit{Cars}, and \textit{Lamps}. We provide 25 $128\times128$ pixel reference images during meta-learning for each 3D object $T$. The reference viewpoints are randomly distributed on a sphere and are oriented towards the target object, and each object is oriented in the canonical coordinate frame. The scenes are lit by a randomly selected environment map \cite{gardner2017learning} and rendered using ray tracing. We use Reptile to meta-learn the initial weights (for each shape category) over 32 unrolled gradient steps. For the MLP $f_\theta$, we use 6 layers with 256 channels each and ReLU nonlinearities, and apply a positional encoding to the input coordinates~\cite{nerf}.

\begin{table}
    \begin{center}
    \input{tables/2d_image_table}
    \end{center}
    \caption{
        Comparison of different initialization methods on an image regression task using the CelebA dataset. We report reconstruction PSNR after two steps of test-time optimization. The meta-learned initialization (\textit{Meta}) significantly outperforms all other initializations. We also report the average number of iterations necessary to match the accuracy of \textit{Meta} after two steps.
    }
    \label{tab:2d_image_mem}
\end{table}

\paragraph{View synthesis for Phototourism~\cite{jin2020phototourism} scenes} 
This dataset consists of thousands of posed tourist photographs of famous landmarks. Our objective is to use these images to create an underlying representation that can be explored and rendered from novel viewpoints with varying lighting conditions. The primary challenge is the diversity of the capture conditions: the photos are taken with different lighting conditions, camera hardware, camera viewpoint, and varying transient objects like people and cars. Each underlying dataset $\mathcal T$ for meta-learning $\theta_0^*$ consists of images of a single landmark (\textit{Trevi}, \textit{Sacre Couer}, or \textit{Brandenburg}); the category is the overall 3D structure of the landmark itself, and the signal is its particular appearance (resulting from the time of day, lighting, weather conditions, etc) within a single photo. If a standard NeRF model is trained directly on this data, it learns a blurry representation of the scene that roughly corresponds to the mean of the environmental conditions. NeRF in the Wild~\cite{martin2020nerfw} explores these shortcomings and proposes extensive architectural modifications to account for the variations. We find that these shortcomings can be addressed to some degree solely with a better initialization and no architectural changes.

We apply meta-learning to the same simple-NeRF model from the ShapeNet experiment. The meta-training dataset for each landmark consists of thousands of images with varying resolution and intrinsic/extrinsic camera parameters. We use Reptile to meta-learn the initial weights (for each landmark) over 64 unrolled gradient steps. At test time, we optimize the simple-NeRF (starting from the initial weights $\theta_0^*$ for that landmark) to reproduce the appearance of a new image, and then render that simple-NeRF from other viewpoints. For the underlying MLP $f_\theta$, we use 6 layers with 256 channels each and ReLU nonlinearities, and apply positional encoding to the input coordinates~\cite{nerf}.

\begin{table}
    \begin{center}
    \input{tables/2d_image_table_confusion}
    \end{center}
    \caption{
       PSNR comparison of four different learned initializations for image regression. Each row corresponds to an initialization meta-learned over a different underlying image dataset. The columns indicate which dataset images are sampled from during testing. 
       The best initialization for each task (bolded) is the one specifically optimized on training images drawn from the same dataset. We observe that initializations transfer better between more similar datasets (\textit{CelebA} and \textit{Imagenette}, both natural images) and poorly between less similar datasets (the frequency spectrum of \textit{Text} images is unlike that of the other categories).
    }
    \label{tab:2d_image_mem_conf}
\end{table}

\subsection{Baselines}

As well as a \textit{Standard} randomly initialized network (Glorot \etal~\cite{glorot2010}), we compare to various other initialization schemes in several of our experimental settings:
\begin{compactitem}
    \item \textit{Mean:} we optimize a network from scratch such that its output matches the mean signal $E_{T\sim \mathcal T}[T]$ from the current class $\mathcal T$.
    \item \textit{Matched:} we optimize a network from scratch such that its output matches the output of a network using the meta-learned initialization for the current class $\mathcal T$.
    \item \textit{Shuffled:} we randomly permute the weights (within each network layer) of the meta-learned initialization $\theta_0^*$ for the current class $\mathcal T$.
\end{compactitem}
Both the \textit{Mean} and \textit{Matched} baselines demonstrate the difference between having a good initialization in \emph{signal} space versus \emph{weight} space---despite \textit{Mean} and \textit{Matched} being initialized so that the loss against a randomly sampled signal will be low, they are a worse starting point for gradient descent than the actual meta-learned initial weights. The \textit{Shuffled} baseline demonstrates that matching the statistical distribution of the meta-learned initial weights is not sufficient for better convergence or generalization. We find that using the Adam~\cite{adam} optimizer performs best for all of the baseline initializations, but that standard stochastic gradient descent works best for the meta-learned initializations (we choose the best optimizer and hyperparameters for each task and initialization using a held-out validation set, see supplement for details).

\begin{table}
    \begin{center}
    \input{tables/ct_table}
    \end{center}
    \caption{
        Comparison of initialization methods on a CT reconstruction task. Each ``view'' consists of 256 parallel rays. The data-dependent prior acquired during meta-learning improves reconstruction quality when fewer views are observed. 
    }
    \label{tab:ct}
\end{table}

\subsection{Faster convergence}

\paragraph{Image regression} 
In Figure~\ref{fig:2d_image_mem_all}, we visualize the network output for a variety of initial weight settings, showing the output images after 0, 1, and 2 gradient steps of test-time optimization. The meta-learned initial weights are optimized to represent face images (CelebA~\cite{liu2015celeba}).
When using the learned initial weights $\theta_0^*$ (\textit{Meta}), the target image is already clearly visible after the very first step. In contrast, the baseline initialization methods take an order of magnitude more iterations to represent the target image to the same accuracy (see Table~\ref{tab:2d_image_mem}). 
The \textit{Mean}, \textit{Matched}, and \textit{Shuffled} baselines perform better than the completely random \textit{Standard} initialization, but still take over ten times as many iterations to reach the same quality as the meta-initialized network can after 2 steps. In particular, this demonstrates that neither matching the image space output nor the statistical distribution of the meta-learned weights is sufficient for achieving a similar speedup. 

\begin{figure}[t]
\begin{center}
   \includegraphics[width=\linewidth]{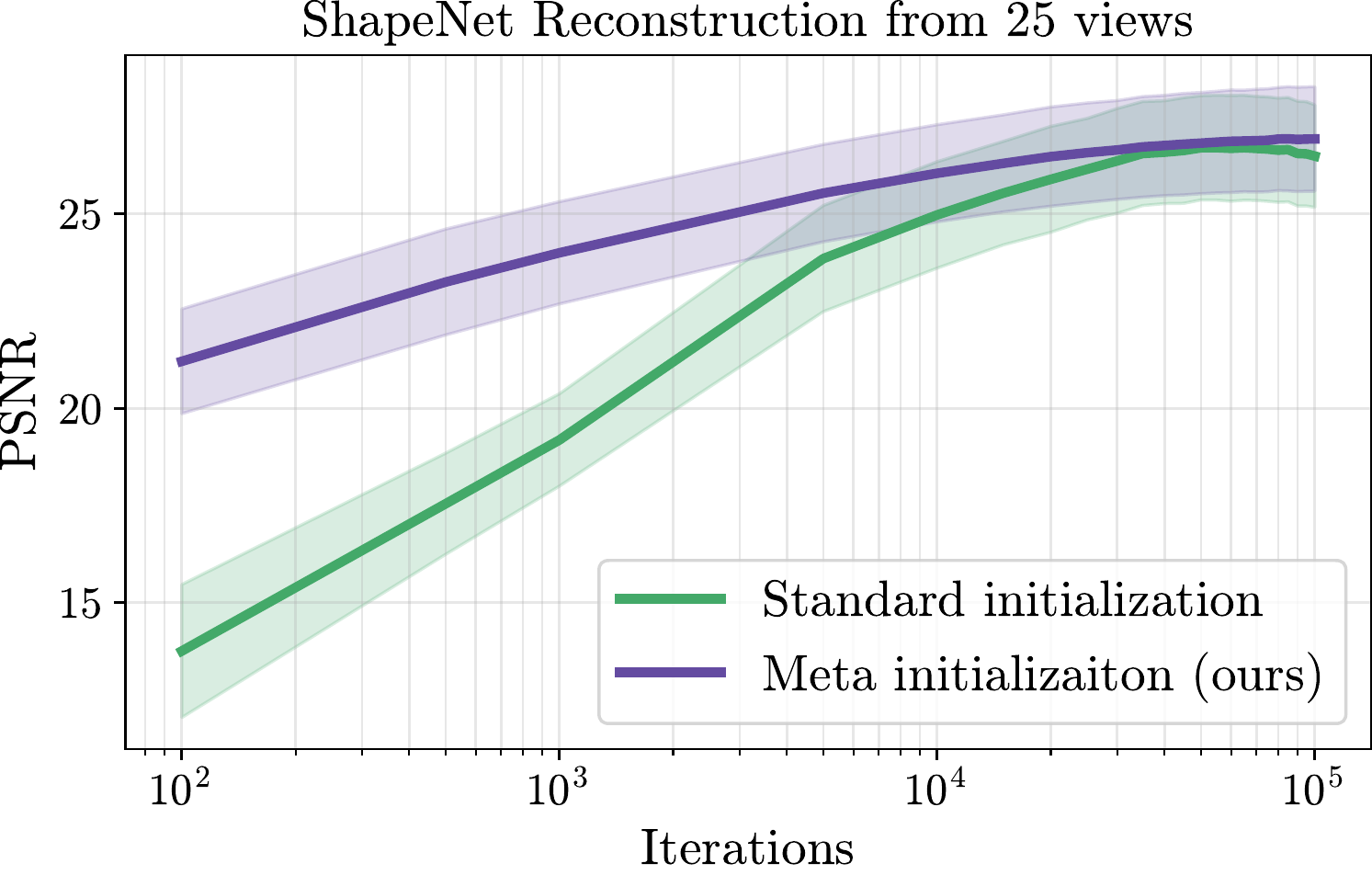}
\end{center}
   \caption{Reconstruction quality over the course of training for models optimized to reconstruct ShapeNet chairs from a set of 25 reference images. The model starting from the meta-learned initial weights outperforms the network using a standard random initialization throughout training.}
\label{fig:shapenet_plot}
\end{figure}

\paragraph{View synthesis for ShapeNet~\cite{shapenet2015} objects} 
In Figure~\ref{fig:shapenet_plot}, we plot the image reconstruction accuracy for a held-out test set of objects from the \textit{Chair} category. During test-time optimization, 25 views are observed. We find that starting from the optimized weights $\theta_0^*$ allows the network to recover the chair more quickly compared to the \textit{Standard} weight initialization. We note that after many steps, both methods end up at a similar quality.

\begin{figure}[t]
\begin{center}
   \includegraphics[width=\linewidth]{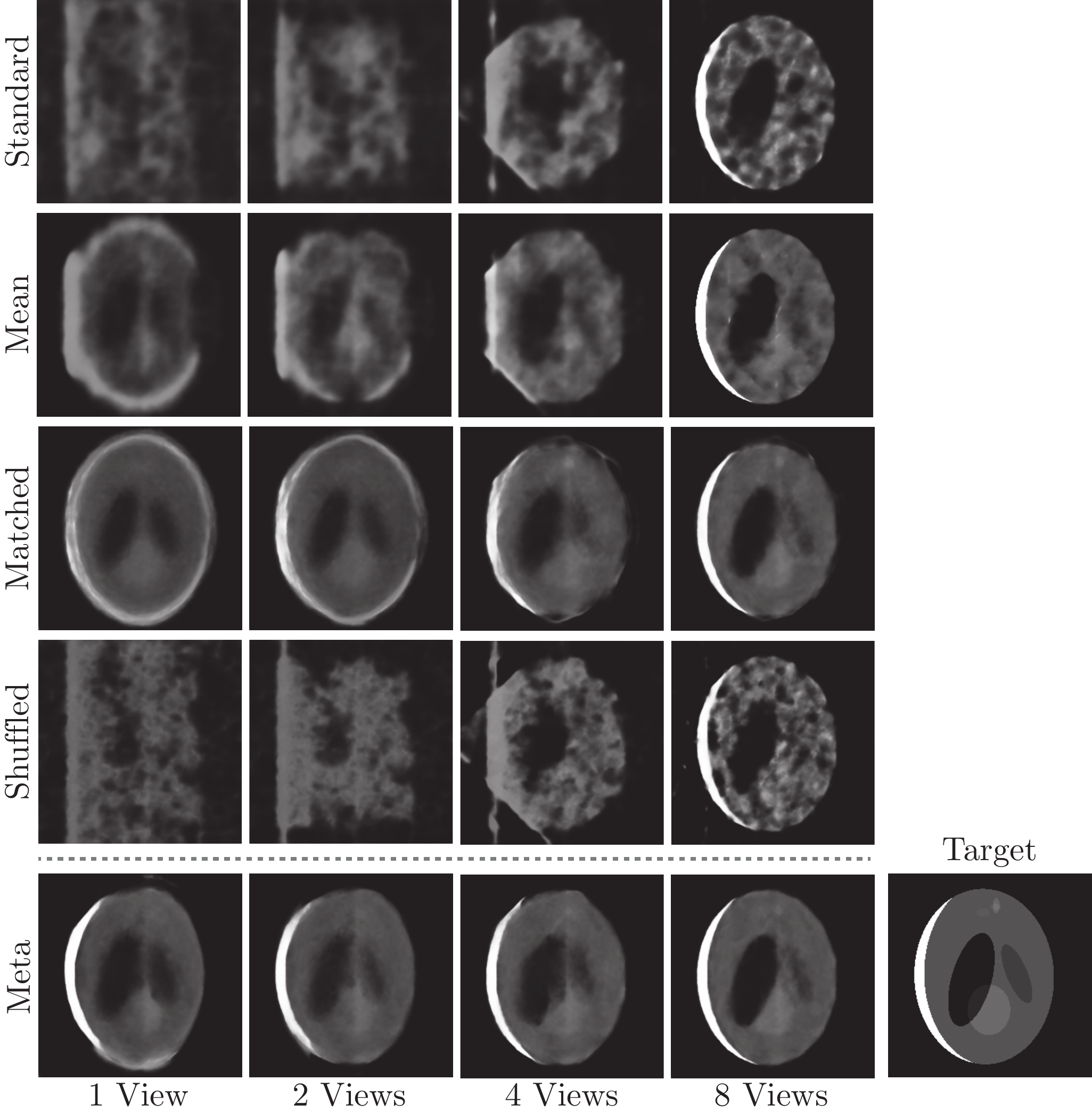}
\end{center}
   \caption{
   \textbf{Sparse Recovery:} Examples of CT reconstructions of a Shepp-Logan phantom from a sparse set of views. The meta-learned initial weights encode a data-dependent prior that improves reconstruction in the limited data regime.
   }
\label{fig:ct_reconstructions}
\end{figure}

\begin{figure*}[t]
\begin{center}
   \includegraphics[width=\linewidth]{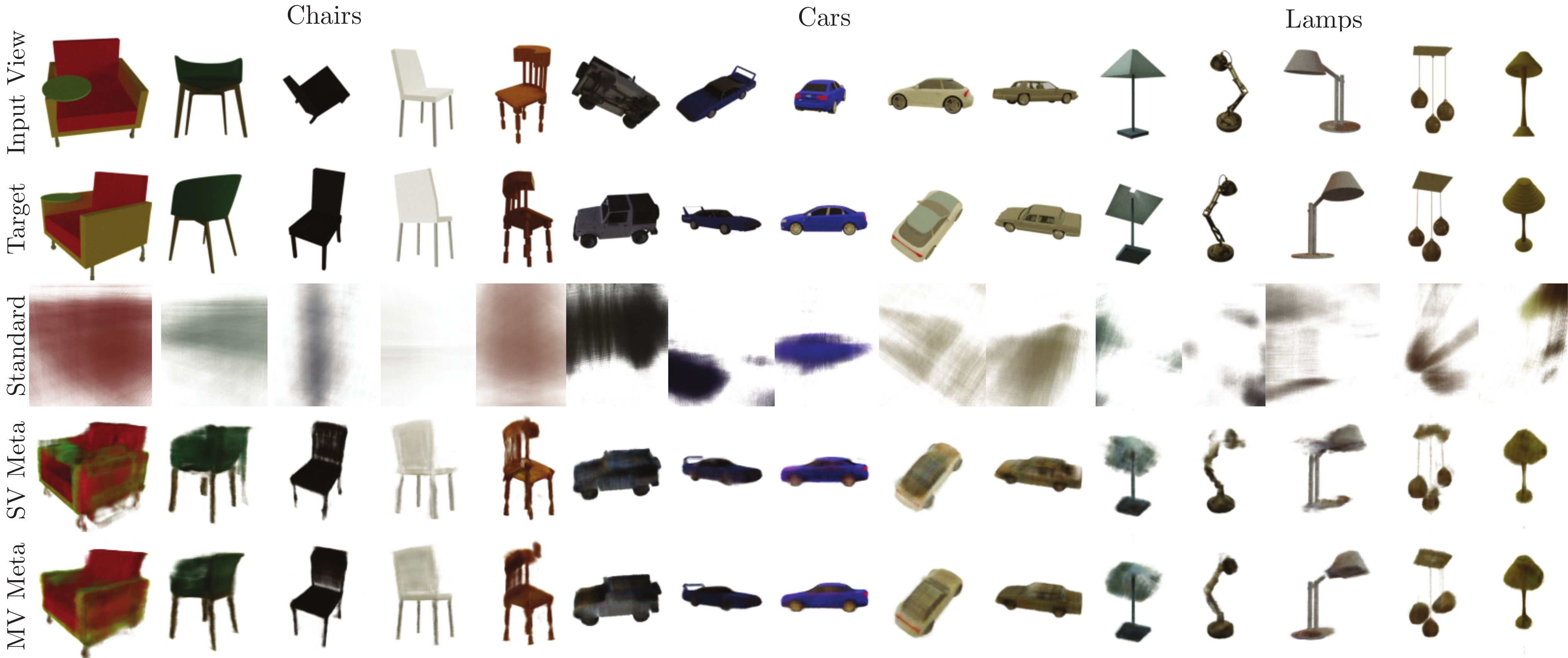}
\end{center}
   \caption{\textbf{Single view reconstructions of ShapeNet~\cite{shapenet2015} objects.} The simple-NeRF formulation relies on multi-view consistency for supervision and therefore fails if naively applied to the task of single view reconstruction, as seen in the \textit{Standard} column. However, if the model is trained starting from meta-learned initial weights, it is able to recover 3D geometry. The \textit{MV Meta} initialization has access to multiple views per object during meta-learning, whereas the \textit{SV Meta} initialization only has access to a single view per object during meta-learning. All methods only receive a single input view during test-time optimization.}
\label{fig:shapenet_sv_results}
\end{figure*}

\subsection{Generalizing from partial observations}

\paragraph{Image regression within a category} 
We perform meta-learning experiments across multiple datasets to determine the extent that the optimized weight initialization acts as a class-specific prior. We compare initializations trained on four different image datasets (\textit{CelebA}, \textit{Imagenette}, \textit{Text}, and \textit{SDF}). Table \ref{tab:2d_image_mem_conf} presents a confusion matrix demonstrating that optimizing the network initialization does in fact induce a dataset-dependent prior, with each learned initialization generalizing best to the same dataset distribution it was trained on.

\paragraph{CT reconstruction from sparse views}
We report the reconstruction quality over a test set of phantoms given varying numbers of views at test time in Table~\ref{tab:ct} and visualize one test example in Figure~\ref{fig:ct_reconstructions}. We observe poor reconstructions from the \textit{Standard} initialization when few views are provided. The meta-learned initializations are consistently able to match the PSNR of \textit{Standard} with half as many views. The \textit{Mean} initialization is generated by training a network to reconstruct the mean of the training phantoms. It is better able to preserve the structure of the phantom compared to \textit{Standard} but still performs worse than the meta-learned initializations.

\begin{figure*}[t]
\begin{center}
   \includegraphics[width=\linewidth]{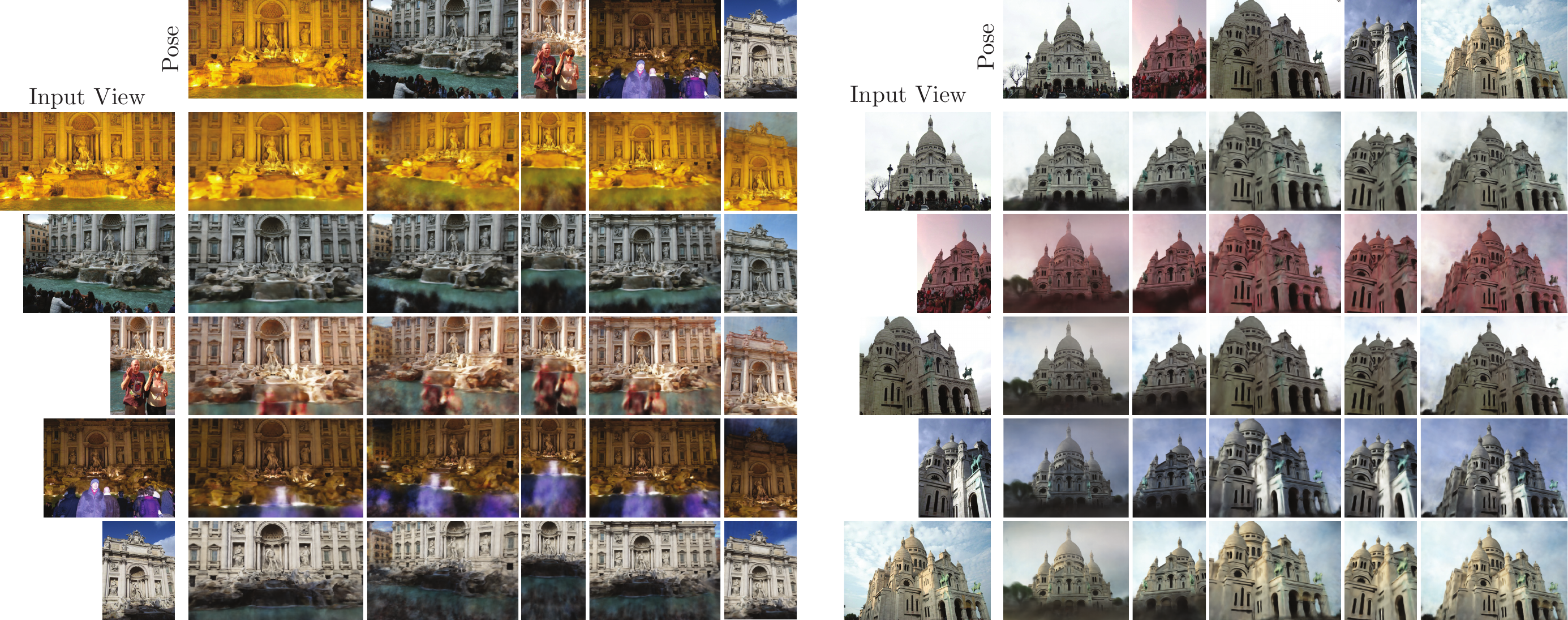}
\end{center}
   \caption{Reconstructions of the Trevi Fountain and Sacre Coeur landmarks from the Phototourism dataset~\cite{jin2020phototourism}. The meta-learning algorithm is run over tourist images taken at different locations and times. During the test-time optimization, the neural representation is trained to recover the input view on the left. The strong prior from the initialization captures the underlying geometry, allowing us to render views from the camera positions of the images in the top row while retaining the appearance of the input view.}
\label{fig:phototourism_results}
\end{figure*}

\paragraph{Single image view synthesis for ShapeNet~\cite{shapenet2015}} \label{sec:shapenet}
A simple-NeRF model with a \textit{Standard} random initialization relies on multi-view consistency to reconstruct the appearance of a 3D object. With only a single view, this na\"ive model is unable to recover any meaningful shape. We find that a learned initialization ``bakes in'' a class-specific shape prior that enables the recovery of 3D geometry (Figure~\ref{fig:shapenet_sv_results}, Table~\ref{tab:shapenet}). We can meta-learn an effective weight initialization for single-view reconstruction by optimizing over a dataset with 25 training views of each object (\textit{MV Meta}). We find that this prior persists even if the meta-training dataset only contains a single reference image per scene (\textit{SV Meta}), meaning that the meta-learning phase has no access to multiview information for any particular object.

\begin{table}
    \begin{center}
    \input{tables/shapenet_table}
    \end{center}
    \caption{
        Metrics for single image ShapeNet reconstructions using a simple-NeRF model. See Figure~\ref{fig:shapenet_sv_results} for image examples and \sref{sec:shapenet} for experimental details.
    }
    \label{tab:shapenet}
\end{table}

\paragraph{View synthesis with appearance transfer for Phototourism~\cite{jin2020phototourism}} 
As described in \sref{sec:tasks}, these images have different camera poses and visual appearance (lighting, sky, etc.) as they are taken by tourists at different times.
Our goal at test time is to explore the landmark from varying camera viewpoints but rendered with the same appearance as in a target photograph. 
In every step of the meta-learning outer loop, we supervise the simple-NeRF model to match the appearance of a random photo of the landmark (with varying pose and appearance).
We find that performing test-time optimization using a single new photograph allows us to render convincing unobserved viewpoints of the scene with the same environmental conditions.

\begin{table}
    \begin{center}

\input{tables/phototourism_table}
    \end{center}
    \caption{Reconstruction results on Phototourism data. Multi-view data with consistent appearance is not available in this dataset, so we optimize on one half of an image and report image metrics on the other half. We compare our Reptile setup (\textit{Meta}) with a standard NeRF network trained on all images of the landmark and then test-time optimized to fit each held-out target image. This is equivalent to training Reptile with one inner loop gradient step.}
    \label{tab:phototourism}
\end{table}

In Figure~\ref{fig:phototourism_results}, we show results for two landmarks. We test-time optimize the meta-learned weights for five target images (shown on the left side of the grid), taking 150 gradient steps for each image. We then render each of the resulting simple-NeRF networks from the five different viewpoints (shown in the row above the grid). The result is an image from the camera position of the corresponding top row image and matching the appearance of the left column image. 

Quantitative evaluation on the Phototourism dataset is difficult as multiple views with the same environmental conditions do not exist. To overcome this, for Table~\ref{tab:phototourism} we optimize and evaluate on the same image, by optimizing to match the appearance of the left half of the image and subsequently evaluating metrics on the right half. For comparison, we train a simple-NeRF model with a standard random initialization from scratch on each landmark, then test-time optimize it to match the left half of each new view before evaluating it on the right half. This is algorithmically equivalent to Reptile with one inner optimization step. We find that unrolling Reptile for 64 inner steps performs better, producing significantly clearer renderings of the landmark. % (see supplemental video).

%------------------------------------------------------------------------
\section{Conclusion}

Our results show that simply modifying a coordinate-based neural representation's initial weight values can guide the network along a significantly better optimization trajectory, without changing the underlying architecture or test-time optimization procedure.
These meta-learned initial weights can result in faster convergence or act as a strong prior for representing signals from a given distribution.
This partially ameliorates a major shortcoming of neural representations (separately optimizing a network for each new signal) 
% separately optimized to encode each new signal---
without limiting their representational power.

There are many additional directions to explore, such as applying more sophisticated meta-learning algorithms or more precisely characterizing the geometry of weight space for these networks. One limitation of our current approach is that it requires a sizable dataset of example signals from a target distribution in order to derive beneficial initial weights.
% As such, it currently cannot be applied to tasks like the one addressed in the original NeRF work (high resolution view synthesis for an arbitrary scene). 
Another shortcoming is that our method still requires some amount of test-time optimization.
% which will usually be slower than a feed forward method such as a hypernetwork.

As the number of use cases for neural representations continues to rapidly expand, we believe this work takes an important step toward understanding the importance of their initial weights and optimization behavior. 

\section{Acknowledgements}

MT is funded by an NSF fellowship and a Berkeley DeepDrive grant. BM is funded by Google through the BAIR Commons Program.
Google University Relations provided a generous donation of GCP compute credits. We thank Ruichao Ren from NVIDIA for donating GPU hardware.

{\small
\bibliographystyle{ieee_fullname}
\bibliography{egbib}
}

\begin{appendices}
\input{latex/supplement}

\end{appendices}

\end{document}

%% file: tables/2d_image_table.tex
\centering
\begin{tabular}{l|c|c}
Init. Method  & 2 Step PSNR $\uparrow$& \# of iters to match $\downarrow$ \\ \hline
Standard      & $10.88$ & $37.92\pm6.31$  \\
Mean          & $14.48$ & $25.59\pm4.57$  \\
Matched     & $13.73$ & $26.32\pm4.17$  \\ 
Shuffled & $16.29$ & $25.80\pm4.02$  \\
\arrayrulecolor[rgb]{0.8,0.8,0.8}\hline
% Reptile       & $25.55$ & $\mathbf{9.86\pm7.42}$  \\ 
Meta          & $\mathbf{30.37}$ & \multicolumn{1}{c}{-} \\
\end{tabular}

%% file: tables/2d_image_table_confusion.tex
\centering
\begin{tabular}{@{}r@{}l|cccc}
\multicolumn{1}{l}{}                                & \multicolumn{1}{c|}{} & \multicolumn{4}{c}{Task}                                       \\
                            &                      & CelebA        & Imagenette    & Text          & SDF            \\ 
\cline{1-6}
\multirow{4}{*}{\rotatebox[origin=c]{90}{Init.}\quad} & CelebA               & \textbf{30.37} & 26.44          & 21.53          & 36.45           \\
                                                    & Imagenette           & 28.51          & \textbf{27.07} & 22.63          & 34.80           \\
                                                    & Text                 & 14.65          & 15.83          & \textbf{27.85} & 23.14           \\
                                                    & SDF                  & 19.80          & 20.05          & 17.23          & \textbf{51.73}  \\
% \cline{2-6}
\end{tabular}

%% file: tables/ct_table.tex
\arrayrulecolor{black}
\begin{tabular}{l|cccc}
\multicolumn{1}{c|}{Init.}  & \multicolumn{4}{c}{PSNR}  \\
\multicolumn{1}{c|}{Method} & \multicolumn{1}{c}{1 Views} & \multicolumn{1}{c}{2 Views} & \multicolumn{1}{c}{4 Views} & 8 View        \\ 
\hline
Standard                    & $13.63$ & $14.15$ & $16.31$ & $21.49$  \\
Mean                        & $14.72$ & $15.39$ & $17.43$ & $25.19$ \\ 
Matched                     & $14.07$ & $15.51$ & $20.25$ & $24.77$ \\ 
Shuffled                    & $13.64$ & $14.17$ & $16.69$ & $22.09$                                      \\ 
\arrayrulecolor[rgb]{0.8,0.8,0.8}\hline
Meta                     & $\mathbf{15.09}$ & $\mathbf{18.70}$ & $\mathbf{22.00}$ & $\mathbf{27.34}$ \\
% MAML                        & $26.04$                 & $21.89$                 & $18.60$                 & $14.80$
% \arrayrulecolor{black}\hline
\end{tabular}

%% file: tables/shapenet_table.tex
\begin{tabular}{l|ccc}
 & \multicolumn{3}{c}{PSNR}      \\
      & Chairs     & Cars   & Lamps    \\ 
\hline
Standard & $12.49$ & $11.45$ & $15.47$  \\
MV Matched & $16.40$ & $22.39$ & $20.79$  \\
MV Shuffled & $10.76$ & $11.30$ & $13.88$  \\
\arrayrulecolor[rgb]{0.8,0.8,0.8}\hline
MV Meta &  $\mathbf{18.85}$  &  $\mathbf{22.80}$   & $\mathbf{22.35}$  \\
SV Meta  &    $16.54$   &    $22.10$     &    $20.95$        
\end{tabular}

%% file: tables/phototourism_table.tex
% \begin{tabular}{l|cc|cc|cc}
%                      & \multicolumn{2}{c|}{Trevi}                     & \multicolumn{2}{c|}{Sacre Coeur} & \multicolumn{2}{c}{Brandenburg Gate}           \\
%                      & PSNR                    & SSIM                 & PSNR           & SSIM            & PSNR                    & SSIM                  \\ 
% \hline
% Standard NeRF TTO & $17.14\pm2.29$          & $0.62\pm0.08$        & $17.59\pm2.53$ & $0.73\pm0.08$   & $17.77\pm2.11$          & $0.75\pm0.06$         \\
% Reptile              & $\mathbf{19.35\pm3.50}$ & $\mathbf{0.72\pm0.09}$ & $\mathbf{19.33\pm2.95}$   & $\mathbf{0.79\pm0.07}$    & $\mathbf{19.11\pm2.88}$ & $\mathbf{0.80\pm0.07}$  \\
% \hline
% \end{tabular}

\begin{tabular}{l|ccc}
 & \multicolumn{3}{c}{PSNR}      \\
                     & Trevi & Sacre Coeur & Brandenburg       \\
\hline
Basic NeRF & $17.14$                & $17.59$   & $17.77$           \\
Meta              & $\mathbf{19.35}$ &  $\mathbf{19.33}$   &     $\mathbf{19.11}$ 
\end{tabular}

%% file: latex/supplement.tex
\section{Implementation details}

We found that modifying the weight initialization for these coordinate-based networks drastically changed their convergence behavior during test-time optimization. As a result, we tuned the optimization method and hyperparameters for each part of each experiment (using held-out validation sets) in order to provide the fairest possible comparison and to not bias the results against the non-meta-learned initializations. For example, we often found that SGD outperformed Adam when doing test-time optimization using meta-learned initializations, but that Adam was significantly better than SGD with a standard random initialization.

All experiments are implemented in JAX~\cite{jax2018github}. Each experiment is trained on either a single NVIDIA V100, 2080 Ti, or 3080 Ti. 
In all cases where the Adam optimizer~\cite{adam} is used, we keep the standard parameter choices for $\beta_1=0.9$, $\beta_2=0.999$, $\epsilon=10^{-8}$. 

\subsection{Image regression}
For this task we use a SIREN~\cite{sitzmann2020siren} architecture ($\omega_0=200$) with 5 layers of 256 channels each. For the randomly initialized \textit{Standard} baseline, we use the specific initialization procedure as proposed in the SIREN paper.

MAML~\cite{maml} is trained for 150K iterations. Each iteration has an outer batch size of 3 target images. The inner batch contains all pixels of the target image. The outer loop uses the Adam optimizer with learning rate of $10^{-5}$. The inner loop performs two steps of gradient descent with a learning rate of $10^{-2}$.

We additionally meta-learn another initialization using Reptile~\cite{reptile}. We use the same learning rates as in MAML but with an outer batch size of 10 target images. We report the Reptile reconstruction accuracy in Table~\ref{tab:supp_image}. We note that Reptile also outperforms the non-meta-learned weights.

During test-time optimization, we use gradient descent with learning rate of $10^{-2}$ when starting from the MAML initial weights. For the baseline methods (\textit{Standard}, \textit{Mean}, \textit{Matched}, \textit{Shuffled}) we used Adam with learning rate of $10^{-4}$, which performed significantly better than than gradient descent.

\subsection{CT reconstruction}

For this task we use an MLP with 5 layers of 256 channels each. The network uses a ReLU activation after each layer with the exception of the last layer, which has a sigmoid activation. Prior to inputting the coordinates into the network, we encode them using random Fourier features sampled from a normal distribution with $\sigma=30$, as was done in Tancik \etal~\cite{tancik2020fourfeat}.

\begin{table}
    \begin{center}
    \input{tables/supp_image_table_ext}
    \end{center}
    \caption{
        Image reconstruction results with meta-learning results for both MAML and Reptile. MAML performs best, but Reptile also outperforms the non-meta-learned weights. 
    }
    \label{tab:supp_image}
\end{table}

Reptile~\cite{reptile} is trained for 100K iterations. Each iteration has an outer batch size of 1. The inner batch contains 20 CT projections, each with 256 measurements, taken from a randomly sampled direction. The outer loop uses the Adam optimizer with learning rate of $5\times10^{-5}$. The inner loop performs 12 inner loop steps of gradient descent with a learning rate of $10^{1}$.

We perform test-time optimization experiments with different numbers of supervision views to compare reconstruction quality. We found that the models are more prone to overfitting when fewer views are provided. We tune the learning rate and number of gradient steps for each initialization method according to a held-out set of 16 validation images. We report all of the test-time optimization hyper-parameters in Table~\ref{tab:supp_ct}.

\subsection{ShapeNet~\cite{shapenet2015} view synthesis}
\label{supp:shapenet}

We use a simplified NeRF~\cite{nerf} model for our view synthesis tasks. This model uses a single network rather than two networks (coarse and fine), and we do not provide view directions as input. The network is an MLP with 6 layers, each with 256 channels and ReLU activations. As in NeRF~\cite{nerf}, we apply a positional encoding to each input coordinate with the form 
\begin{equation}
    \bigcup_{i=0}^N \left\{ \cos\left(2^{fi/N}x\right), \sin\left(2^{fi/N}x\right) \right\}\,,
\end{equation}
with $N=20$ encodings and log-max frequency $f=8$.
% \todo{fix: [$sin(x * 2^i), cos(x * 2^i)$] for i in linspace(0,8,20)}. 
We accumulate 128 samples per ray for rendering.

Reptile is trained for 100K iterations with an outer batch size of 1. The inner loop step optimizes over a batch of 128 rays. We perform 32 inner loop steps for every outer loop step. The outer loop uses the Adam optimizer with learning rate $5\times10^{-4}$ for the \textit{Chairs} scenes and $5\times10^{-5}$ for the \textit{Lamps} and \textit{Cars} scenes.

The test-time optimization parameters vary depending on the scene and the number of views available during meta-learning. Each experiment uses an inner batch of 64 rays. The \textit{Shuffled} and \textit{Matched} initializations are computed based on the \textit{MV Meta} weights. For the 25 view chair reconstruction, we use stochastic gradient descent with a learning rate of $10^{-1}$ for the Reptile initialization; for the standard initialization, we use Adam with a learning rate of $10^{-4}$. The test-time hyper parameters for the single view experiments are listed in Table~\ref{tab:supp_shapenet}. 

\begin{table*}
    \begin{center}
    \input{tables/supp_ct_hyperparams}
    \end{center}
    \caption{
        Hyper-parameters for CT test-time optimization. Each value is tuned on a held-out validation set. 
    }
    \label{tab:supp_ct}
\end{table*}

\begin{table*}
    \begin{center}
    \input{tables/supp_shapenet_hyperparams}
    \end{center}
    \caption{
        Hyper-parameters for ShapeNet test-time optimization from a single view. Each value is tuned on a held-out validation set. 
    }
    \label{tab:supp_shapenet}
\end{table*}

\subsection{Phototourism~\cite{jin2020phototourism} view synthesis}

We use the same architecture as described in \sref{supp:shapenet}. Reptile is trained for 150K iterations with an outer batch size of 1. The inner loop step optimizes over a batch of 64 rays, with 128 volume rendering samples per ray. The outer loop uses the Adam optimizer with a learning rate of $5\time10^{-4}$. We train with 64 inner loop steps using gradient descent with a learning rate of $10$. We compare to \textit{Basic NeRF} which has the same setup, but only one inner step. For \textit{Basic NeRF} we train \textit{Trevi} for 60K iterations, \textit{Brandenburg} for 100K iterations, and \textit{Sacre Coeur} for 200K iterations. To transfer the appearance of a new photo during test-time optimization, we take 150 gradient steps with a learning rate of $10$. 

\section{Weight space interpolation}

\begin{figure*}[t]
\begin{center}
   \includegraphics[width=\linewidth]{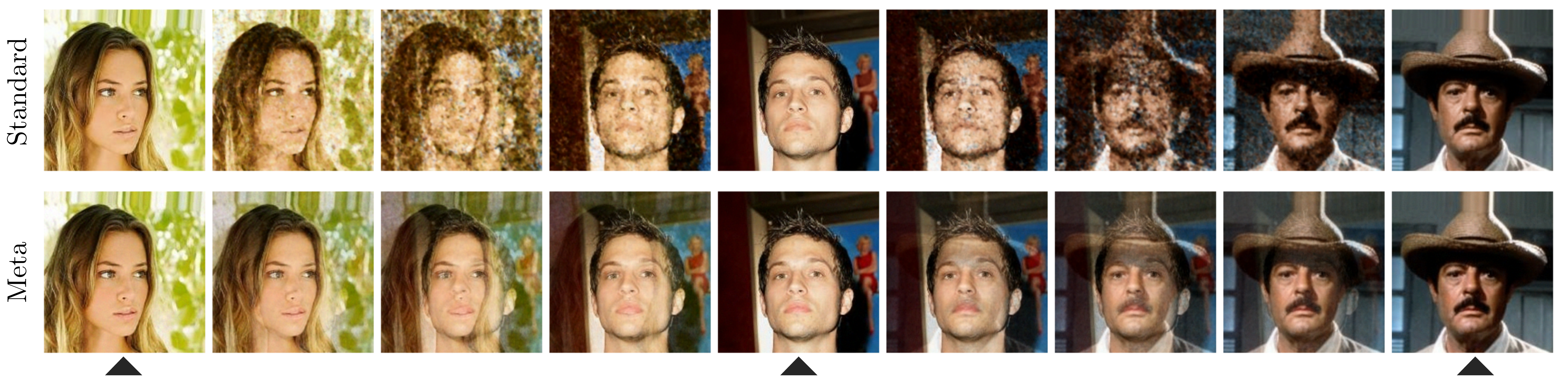}
\end{center}
   \caption{Weight space interpolation for networks optimized to represent 2D images. We use test-time optimization to fit network weights for three different images (denoted with arrows), then linearly interpolate between those weight values and visualize the resulting outputs. When test-time optimization begins from a standard random initialization (\textit{Standard}, top), weight space interpolation produces displeasing artifacts, but when it begins from a meta-learned initialization (\textit{Meta}, bottom) the resulting outputs maintain an image-like appearance.}
\label{fig:img_interp}
\end{figure*}

\begin{figure*}[t]
\begin{center}
   \includegraphics[width=\linewidth]{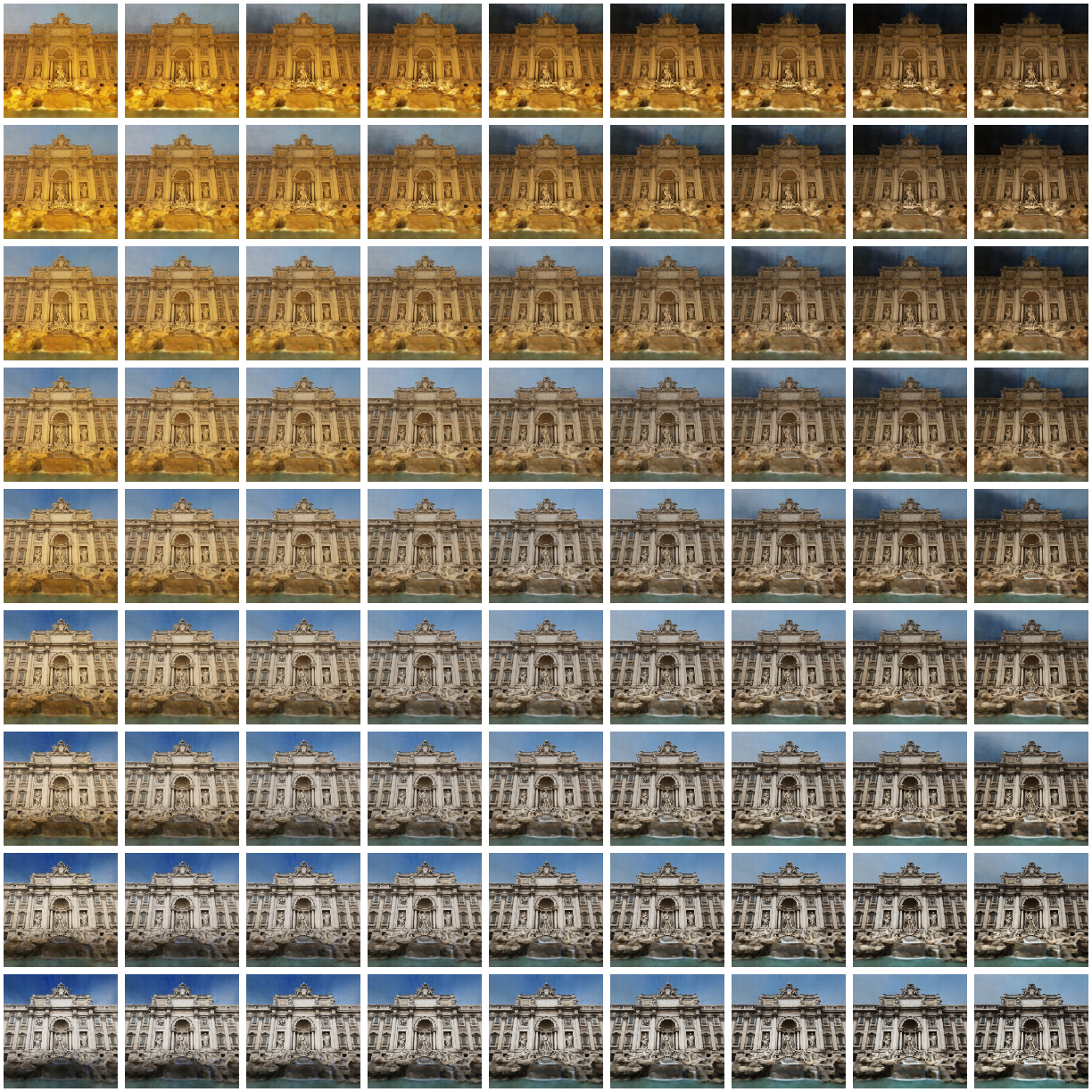}
\end{center}
   \caption{Appearance interpolation on the Trevi Fountain scene from the Phototourism dataset~\cite{jin2020phototourism}. We render the scene from a single fixed camera pose. Each corner of the grid represents a NeRF network that has been test-time optimized to match the appearance of a single image (starting from the meta-learned initial weights $\theta_0^*$). We linearly interpolate between these networks in weight space to render the grid of images shown here. The image in the center is produced by directly rendering the meta-learned initial weights (with no test-time optimziation), representing an ``average'' appearance for the scene. Please see the supplemental video for an animated version of this figure with a moving camera path.}
\label{fig:trevi_interp}
\end{figure*}

We find that linearly interpolating between networks in weight space produces meaningful outputs when using meta-learned weights. Figure~\ref{fig:img_interp} shows interpolation between networks trained to represent images, and Figure~\ref{fig:trevi_interp} shows interpolations between networks that are trained to reconstruct a Phototourism landmark.

% \section{Acknowledgements}

% MT is funded by an NSF fellowship and a Berkeley DeepDrive grant. BM is funded by Google through the BAIR Commons Program.
% Google University Relations provided a generous donation of GCP compute credits. We thank Ruichao Ren from NVIDIA for donating GPU hardware.

%% file: tables/supp_image_table_ext.tex
\centering
\begin{tabular}{l|c|c}
Init. Method  & 2 Step PSNR $\uparrow$& \# of iters to match $\downarrow$ \\ \hline
Standard      & $10.88$ & $37.92\pm6.31$  \\
Mean          & $14.48$ & $25.59\pm4.57$  \\
Matched     & $13.73$ & $26.32\pm4.17$  \\ 
Shuffled & $16.29$ & $25.80\pm4.02$  \\
\arrayrulecolor[rgb]{0.8,0.8,0.8}\hline
Reptile       & $25.55$ & $\mathbf{9.86\pm7.42}$  \\ 
MAML          & $\mathbf{30.37}$ & \multicolumn{1}{c}{-} \\
\end{tabular}

%% file: tables/supp_ct_hyperparams.tex
\begin{tabular}{l|cc|cccc}
 &  \multicolumn{2}{c|}{Opt. method} &  \multicolumn{4}{c}{Number of steps}      \\
     & Adam & LR &  \multicolumn{1}{c}{1 View} & \multicolumn{1}{c}{2 Views} & \multicolumn{1}{c}{4 Views} & \multicolumn{1}{c}{8 Views} \\ 
\hline
Standard & \cmark & $5\times10^{-4}$ & $50$ & $100$ & $250$ & $1000$ \\
Mean & \cmark & $10^{-5}$ & $25$ & $50$ & $100$ & $1000$ \\
Matched & \cmark & $5\times10^{-4}$ & $50$ & $100$ & $500$ & $1000$ \\
Shuffled & \cmark & $5\times10^{-4}$ & $50$ & $100$ & $250$ & $1000$ \\
Meta  &  \xmark   &    $10^{1}$     & $50$ & $100$ & $1000$ & $1000$
\end{tabular}

%% file: tables/supp_shapenet_hyperparams.tex
\begin{tabular}{l|ccc|ccc|ccc}
 & \multicolumn{3}{c}{Chairs} &  \multicolumn{3}{c}{Cars}&  \multicolumn{3}{c}{Lamps}     \\
     & Adam & LR & Steps& Adam & LR & Steps& Adam & LR & Steps \\ 
\hline
Standard & \cmark & $10^{-5}$ & $1000$ & \cmark & $5\times10^{-5}$ & $2000$ & \cmark & $5\times10^{-5}$ & $2000$\\
Matched & \cmark & $10^{-4}$ & $2000$ & \cmark & $5\times10^{-5}$ & $2000$ & \cmark & $5\times10^{-5}$ & $2000$\\
Shuffled & \cmark & $10^{-4}$ & $2000$ & \cmark & $5\times10^{-5}$ & $2000$ & \cmark & $5\times10^{-5}$ & $2000$\\
MV Meta  &  \xmark & $10^{-1}$ & $1000$ & \xmark & $5\times10^{-1}$ & $2000$ &  \xmark & $5\times10^{-1}$ & $2000$\\
SV Meta  &  \xmark & $5\times10^{-1}$ & $1000$ &  \xmark  & $5\times10^{-1}$ & $2000$ &  \xmark & $5\times10^{-1}$ & $2000$\\
\end{tabular}